\documentclass[11pt,a4paper]{article}
\usepackage[hyperref]{emnlp2020}

\usepackage{times}
\usepackage{latexsym}
\usepackage{url}
\usepackage{graphicx}
\usepackage{amsmath}
\usepackage{amssymb}
\usepackage{subfigure}
\usepackage{caption}
\usepackage{algorithm}
\usepackage{algorithmicx}
\usepackage{algpseudocode}
\usepackage{multicol}
\usepackage{multirow}
\usepackage{booktabs}
\usepackage{listings}
\usepackage{array}
\usepackage{xspace}
\usepackage{bm}
\usepackage{comment}
\graphicspath{ {./images/} }
\usepackage{lipsum}

\usepackage{microtype}

\aclfinalcopy 
\usepackage{color, colortbl}
\definecolor{GREEN_YL_1}{rgb}{0.4 ,0.6 , 0.8}
\definecolor{GREEN_YL_2}{rgb}{0.48500000000000004 , 0.6666666666666666 , 0.8333333333333334}
\definecolor{GREEN_YL_3}{rgb}{0.5700000000000001 , 0.7333333333333333 , 0.8666666666666667}
\definecolor{GREEN_YL_4}{rgb}{0.655 , 0.8 , 0.9}
\definecolor{GREEN_YL_5}{rgb}{0.74 , 0.8666666666666667 , 0.9333333333333333}
\definecolor{GREEN_YL_6}{rgb}{0.8250000000000001 , 0.9333333333333333 , 0.9666666666666667}
\definecolor{GREEN_YL_7}{rgb}{0.91 , 1.0 , 1.0}

\newcommand\NAME{Coke}
\newcommand\Coke{\NAME\xspace}
\newcommand\KADAPTER{K-A\textsc{dapter}\xspace}
\newcommand\BASESIZE{$_{\small \textsc{Base}}$\xspace}
\newcommand\LARGESIZE{$_{\small \textsc{Large}}$\xspace}
\newcommand\CokeBERTBASE{\NAME$^{\small \textsc{BERT}}_{\small \textsc{Base}}$\xspace}

\newcommand\CokeRoBERTaBASE{\NAME$^{\small \textsc{RoBERTa}}_{\small \textsc{Base}}$\xspace}
\newcommand\CokeRoBERTaLARGE{\NAME$^{\small \textsc{RoBERTa}}_{\small \textsc{Large}}$\xspace}

\title{CokeBERT: Contextual Knowledge Selection and Embedding towards \\ Enhanced Pre-Trained Language Models}

\author{
Yusheng Su$^{1}\thanks{\quad indicates equal contribution}$, Xu Han$^{1*}$, Zhengyan Zhang$^{1}$, Yankai Lin$^{2}$,\\ \textbf{Peng Li}$^{2}$, \textbf{Zhiyuan Liu}$^{1}$, \textbf{Jie Zhou}$^{2}$, \textbf{Maosong Sun}$^{1}$\\
$^{1}$Department of Computer Science and Technology, Tsinghua University, Beijing, China\\
Institute for Artificial Intelligence, Tsinghua University, Beijing, China\\
State Key Lab on Intelligent Technology and Systems, Tsinghua University, Beijing, China\\
$^{2}$Pattern Recognition Center, WeChat AI, Tencent Inc.\\
{\tt\{suys19,hanxu17\}@mails.tsinghua.edu.cn,}\\
}

\begin{document}
\maketitle
\begin{abstract}
Several recent efforts have been devoted to enhancing pre-trained language models (PLMs) by utilizing extra heterogeneous knowledge in knowledge graphs (KGs), and achieved consistent improvements on various knowledge-driven NLP tasks. However, most of these knowledge-enhanced PLMs embed static sub-graphs of KGs (``knowledge context''), regardless of that the knowledge required by PLMs may change dynamically according to specific text (``textual context''). In this paper, we propose a novel framework named \Coke to dynamically select contextual knowledge and embed knowledge context according to textual context for PLMs, which can avoid the effect of redundant and ambiguous knowledge in KGs that cannot match the input text. Our experimental results show that \Coke outperforms various baselines on typical knowledge-driven NLP tasks, indicating the effectiveness of utilizing dynamic knowledge context for language understanding. Besides the performance improvements, the dynamically selected knowledge in \Coke can describe the semantics of text-related knowledge in a more interpretable form than the conventional PLMs. Our source code and datasets will be available to provide more details for \Coke.
\end{abstract}

\section{Introduction}
\label{Introduction}

Pre-trained language models (PLMs) such as BERT~\cite{devlin2018bert} and RoBERTa~\cite{liu2019RoBERTa} have achieved state-of-the-art performance on a wide range of natural language processing (NLP) tasks. As some research~\cite{poerner2019bert} suggests that these PLMs still struggle to learn factual knowledge, intensive recent efforts~\cite{lauscher2019LIBERT,levine2019SenseBERT,ERNIE-baidu,wang2019kepler,zhang2019ernie,Peters2019KnowledgeEC,he2019KRL,Liu2019KBERTEL} have therefore been devoted to leveraging rich heterogeneous knowledge in knowledge graphs (KGs) to enhance PLMs. 

\begin{figure}[t]
    \centering
    \includegraphics[width = 1.0\linewidth]{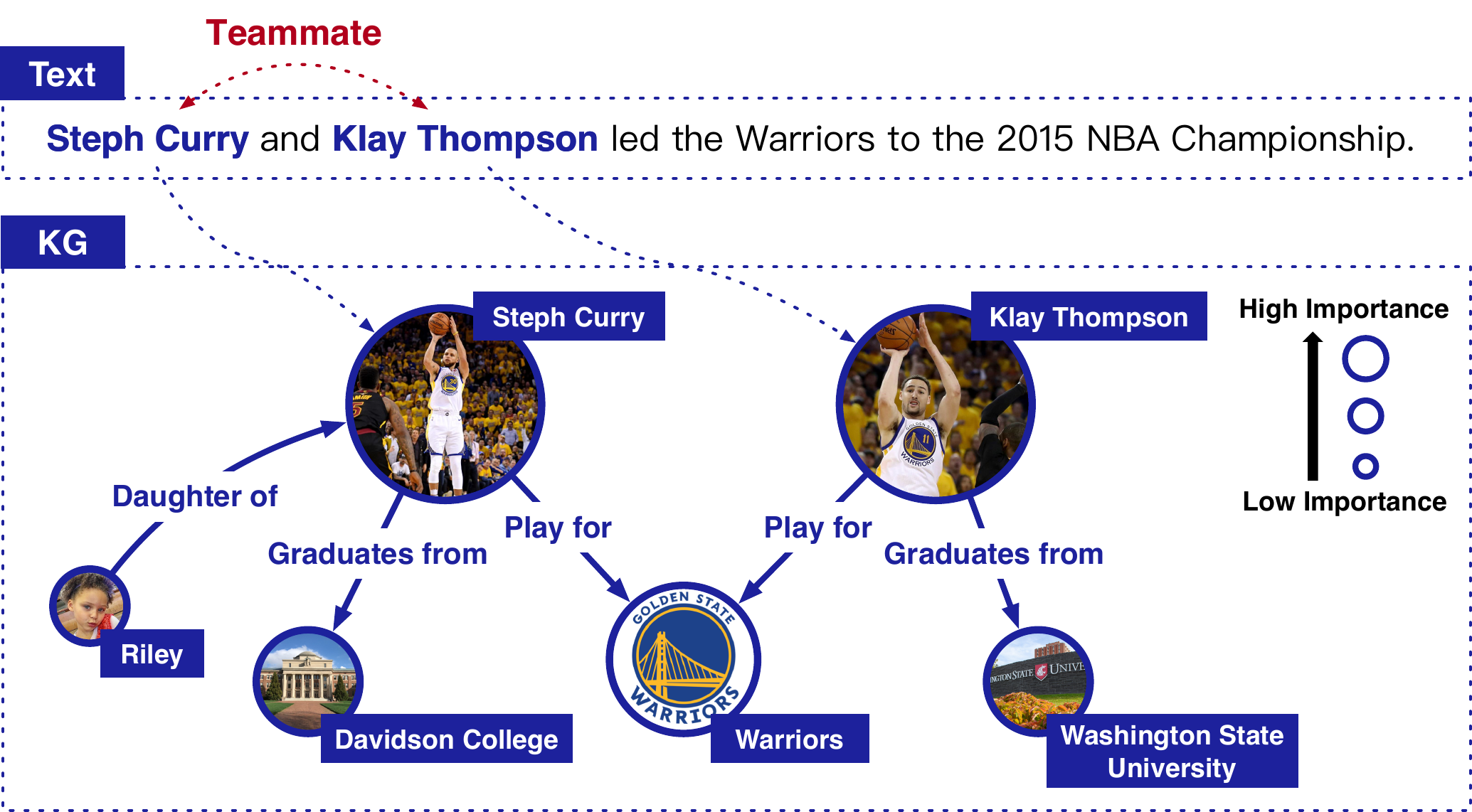}
    \caption{The example of capturing knowledge context from a KG and incorporating them for language understanding. Different sizes of circles express different entity importance for understanding the given sentence.}
    \label{fig:running-example}
\end{figure}

An ideal process for injecting factual knowledge into PLMs is to first identify mentioned entities\footnote{Those words or phrases in the text corresponding to certain entities in KGs are often named ``entity mentions''.} in the input text (``textual context''), then dynamically select sub-graphs (``knowledge context'') centered on these mentioned entities from KGs, and finally embed the selected knowledge context for PLMs. Intuitively, knowledge context contributes to better language understanding on the one hand, serving as an effective complementarity to textual context. For example, given two entities {\it Steph Curry} and {\it Klay Thompson} in Figure~\ref{fig:running-example}, we can infer that they play for the same basketball team, which is not explicitly described in the given sentence. On the other hand, not all knowledge in KGs is relevant to textual context, e.g., the fact ({\it Riley}, {\tt Daughter of}, {\it Steph Curry}) has no positive effect on understanding the given sentence. 

We argue that it is meaningful to dynamically select appropriate knowledge context that can match specific textual context for enhancing PLMs. However, most knowledge context utilized in existing knowledge-enhanced PLMs is not highly matching textual context: (1) ERNIE~\cite{zhang2019ernie} just uses entities mentioned in the text as knowledge context and only injects the embeddings of these entities into PLMs, ignoring informative neighbors in KGs; (2) KnowBert~\cite{Peters2019KnowledgeEC}, K-BERT~\cite{Liu2019KBERTEL} and \KADAPTER~\cite{Wang2020KAdapterIK} consider more information as knowledge context than ERNIE (e.g, entity properties in KGs), yet their knowledge context is still static and cannot dynamically change according to textual context. As we mentioned before, not all information in static knowledge context can match textual context, and the knowledge interfere with redundant and ambiguous information may interfere understanding semantics. Hence, how to dynamically select and embed knowledge context according to textual context for PLMs still remains a challenge.

To alleviate the issue, we propose a novel framework named \Coke to dynamically select knowledge context matching textual context and embed the dynamic context for enhancing PLMs:
(1) For \textbf{dynamically selecting knowledge context}, according to textual context, we propose a novel semantic-driven graph neural network (S-GNN). Given an entity mentioned in textual context, S-GNN leverages an attention mechanism to filter out irrelevant KG information by assigning scores to neighbors ($1$-hop, $2$-hop, etc) and relations between entities based on textual context. The score can weigh how much the information in KGs matches textual context and help \Coke dynamically select an appropriate sub-graph as the knowledge context of the given entity mention. (2) For \textbf{dynamically embedding knowledge context}, given a mentioned entity, S-GNN computes its representation conditioned on both its pre-trained entity embedding and the information aggregated from the selected contextual sub-graph in a recursive way, making \Coke be aware of both global and local KG information and grasp the text-related information. (3) By \textbf{fusing the embeddings of dynamic knowledge context} for PLMs with specific training and adaption strategies, \Coke improves language understanding and benefits for downstream applications.

Following existing work, we conduct experiments on four datasets for two typical knowledge-driven tasks, i.e., entity typing and relation classification. The experimental results show that \Coke outperforms various baselines, indicating the effectiveness of dynamically selecting and embedding knowledge context for PLMs. Moreover, some qualitative analyses also suggest that, as compared with the state-of-the-art knowledge-enhanced PLMs, our model not only achieves competitive results but also provides a more interpretable approach to describing specific words based on their dynamic knowledge context.

\section{Related Work}
\label{Related Work}

Intuitively, two types of context are involved in language understanding: (1) the semantic information of the text ({\em textual context}), and (2) the factual knowledge related to the text ({\em knowledge context}). The typical PLMs focus on capturing information from the textual context, like ELMO~\cite{peters2018deep}, GPT~\cite{radford2018improving}, BERT~\cite{devlin2018bert}, XLNET~\cite{yang2019xlnet}, and RoBERTa~\cite{liu2019RoBERTa}. In order to enable PLMs to better understand the knowledge context, intensive efforts have been devoted to injecting various factual knowledge of KGs into PLMs. ERNIE~\cite{zhang2019ernie} links entity mentions in textual context to their corresponding entities in KGs and then inject the pre-trained embeddings of the corresponding entities into PLMs. Although ERNIE has shown the feasibility and effectiveness of fusing knowledge embeddings for enhancing PLMs, it still doees not consider the informative neighbors of entities.

To this end, various models have been proposed to further incorporate a wider range of knowledge information. KnowBert~\cite{Peters2019KnowledgeEC} and KRL~\cite{he2019KRL} employ attention mechanisms to learn more informative entity embeddings based on the entity-related sub-graphs. Nevertheless, the computation of entity embeddings is independent of textual context. K-BERT~\cite{Liu2019KBERTEL} heuristically converts textual context and entity-related sub-graphs into united input sequences, and leverages a Transformer~\cite{vaswani2017transformer} with a specially designed attention mechanism to encode the sequences. Unfortunately, it is not trivial for the heuristic method in K-BERT to convert the second or higher order neighbors related to textual context into a sequence without losing graph structure information. \KADAPTER~\cite{Wang2020KAdapterIK} proposes variant frameworks to inject factual knowledge in different domains, yet still suffers from the similar issue like K-BERT. Although most existing knowledge-enhanced PLMs are aware of utilizing both textual context and knowledge context, their knowledge context cannot change with textual context, like ERNIE using single entities, KRL and KnowBert embedding sub-graphs independently of textual context, K-BERT and \KADAPTER using fixed sub-graphs. In contrast, our proposed \Coke model can leverage dynamic sub-graphs of arbitrary size as knowledge context according to textual context. 
 

There are also several PLM methods for capturing knowledge from only textual context. SpanBERT~\cite{joshi2019spanbert} and ERNIE~1.0-Baidu~\cite{ERNIE-baidu} propose to predict masked variable-length spans or entity mentions to encourage PLMs to learn multi-token phrases.
WKLM~\cite{Xiong2019PretrainedEW} is trained to distinguish whether an entity mention has been replaced with the name of other entities having the same type to learn entity types.
LIBERT~\cite{lauscher2019LIBERT} and SenseBERT~\cite{levine2019SenseBERT} extend PLMs to predict word relations (e.g., synonym and hyponym-hypernym) and word-supersense respectively to inject lexical-semantic knowledge. Moreover, there are also efforts on continual knowledge infusion~\cite{sun2020ERNIE2,Wang2020KAdapterIK}. Although these models do not use extra knowledge context to understand factual knowledge, they are complementary to our work and can be used together towards better PLMs.


\begin{figure*}[t]
\centering  
\subfigure{
\label{methodology_a}
\includegraphics[width = 1.0\linewidth]{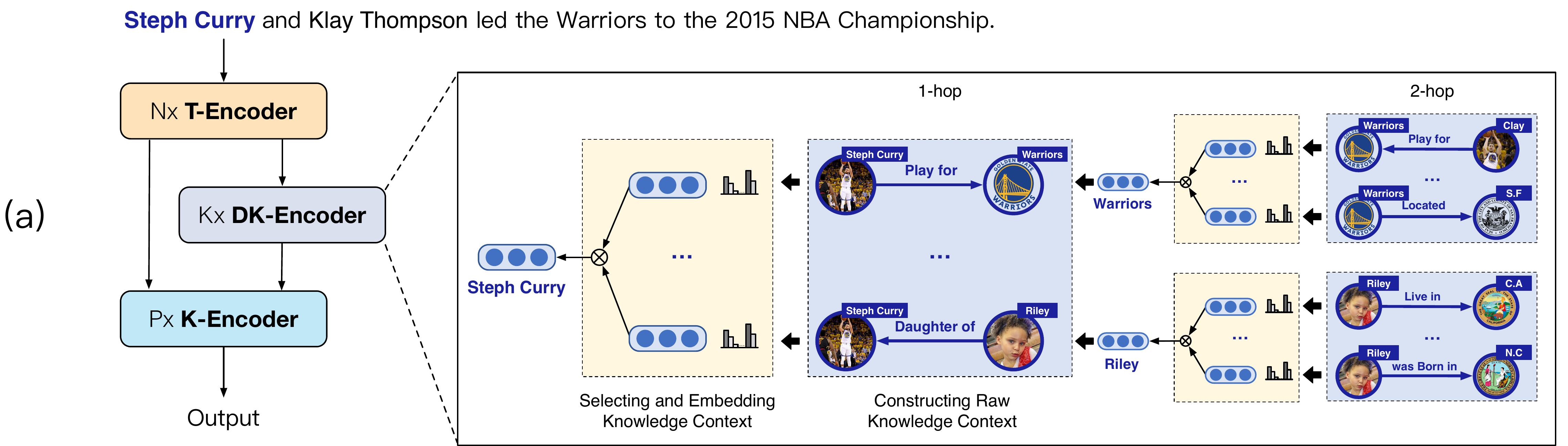}
}
\\
\subfigure{
\label{methodology_b}
\includegraphics[width = 1.0\linewidth]{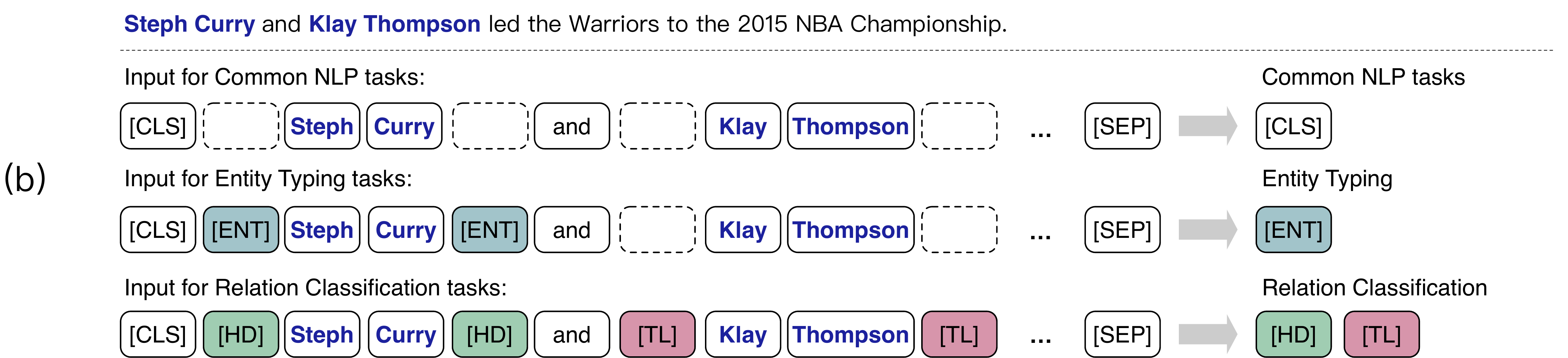}}
\caption{(a)~The upper part is the overall framework of \Coke and illustrates how to generate entity representations. (b)~The lower part is the example of inserting special tokens to the input sequence for specific tasks during fine-tuning.}
\label{fig:methodology}
\end{figure*}

\section{Methodology}
\label{Methodology}

As shown in Figure~\ref{fig:methodology}, \Coke consists of three modules:

(1) {\bf Text Encoder} computes embeddings for the input text, i.e. textual context;

(2) {\bf Dynamic Knowledge Context Encoder} first dynamically selects knowledge context according to textual context, and then computes contextual knowledge embeddings conditioned on both textual context and KG context;

(3) {\bf Knowledge Fusion Encoder} fuses both textual context and dynamic knowledge context embeddings for better language understanding. In this section, we will first give the notations and then present the three modules in details.

\subsection{Notations}
\label{Notations}

A KG is denoted by $\mathcal{G}=\{(h,r,t)|h,t\in\mathcal{E},r\in\mathcal{R}\}$, where $\mathcal{E}$ and $\mathcal{R}$ are the set of entities and relations respectively. For each fact $(h,r,t) \in \mathcal{G}$, it indicates that there is a relation $r$ between the head entity $h$ and the tail entity $t$. Given a token sequence $S=\{w_{j}\}_{j=1}^{N}$ of the length $N$, some tokens in the sequence may correspond to certain entities in $\mathcal{E}$, we name these tokens ``entity mentions'' and denote their mentioned entities in KGs as $\{e_j\}_{j=1}^{M}$, where $M$ is the number of mentioned entities\footnote{Typically, $M \ne N$ as an entity may correspond to multiple different tokens. In this work, we use the toolkit TAGME to identify the mentioned entities.}. 

\subsection{Text Encoder}
\label{sec:textual-encoder}

Similar to existing knowldege-enhanced PLMs, \Coke leverages a $L$-layer bidirectional Transformer encoder~\cite{vaswani2017transformer,devlin2018bert} to embed the input text~(tokens) $S=\{w_j\}_{j=1}^N$ and obtain its textual context representations, which is denoted as $\texttt{T-Encoder}(\cdot)$,
\begin{equation}
{\{\mathbf{w}_{j}\}_{j=1}^{N}} = \texttt{T-Encoder}({\{w_{j}\}_{j=1}^{N}}).
\label{eqn:textual-encoder}
\end{equation}
As $\texttt{T-Encoder}(\cdot)$ is the same as that used in BERT, we refer the readers to the original paper~\cite{devlin2018bert} for more details. 


\subsection{Dynamic Knowledge Context Encoder}
\label{sec:DCAK-Encoder}

\paragraph{Constructing Raw Knowledge Context}

As KGs are often in a large scale, we first construct raw knowledge context for computational efficiency. Then we dynamically select and embed appropriate knowledge context that can match the textual context. Specifically, given a mentioned entity $m \in \mathcal{E}$ mentioned by the input text $S=\{w_j\}_{j=1}^N$, we define its raw knowledge context $\mathcal{G}_m$ as a sub-graph of $\mathcal{G}$ centered in $m$. The entities of $\mathcal{G}_m$ are at most $K$-hops away from $m$. Formally, we define the $0$-hop away entity set as $\mathcal{E}^0_m=\{m\}$. Then the $i$-hop away entity set $\mathcal{E}^i_m$ can be defined recursively as
\begin{equation}
\begin{aligned}
    \overrightarrow{\mathcal{E}}^i_m &=\left\{t\left|
        \begin{array}{c}
            h\in \mathcal{E}^{i-1}_m \land t \notin \bigcup_{j=0}^{i-1}{\mathcal{E}}^j_m, 
            (h,r,t)\in\mathcal{G}
        \end{array}
        \right.\right\},\\
    \overleftarrow{\mathcal{E}}^i_m &=\left\{h\left|
        \begin{array}{c}
            t\in \mathcal{E}^{i-1}_m  \land  h\notin \bigcup_{j=0}^{i-1}{\mathcal{E}}^j_m, 
            (h,r,t)\in\mathcal{G}
        \end{array}
        \right.\right\},\\
    \mathcal{E}^i_m&=\overrightarrow{\mathcal{E}}^i_m\cup\overleftarrow{\mathcal{E}}^i_m.
\end{aligned}
\end{equation}
Intuitively, all entities in $\mathcal{E}^i_m$ (both head or tail entities) only have relations to the entities in $\mathcal{E}^{i-1}_m$. Then, the raw knowledge context $\mathcal{G}_m$ and its entity set $\mathcal{E}_m$ can be defined as
\begin{equation}
\begin{aligned}
    \mathcal{E}_m &= \bigcup_{i=0}^{K}{\mathcal{E}}_m^i\\
    \mathcal{G}_m &=\left\{(h,r,t)\left|
        \begin{array}{c}
            h \in \mathcal{E}_m \land t \in \mathcal{E}_m,\\
            (h,r,t) \in \mathcal{G}
        \end{array}
        \right.\right\}.
\end{aligned}
\end{equation}


\paragraph{Selecting and Embedding Knowledge Context}

To dynamically select informative features in $\mathcal{G}_m$ and embed these features for PLMs, we propose a semantic-driven graph neural network (S-GNN). For each entity in $\mathcal{G}_m$, i.e., $e \in \mathcal{E}_m$, we initialize its input features for S-GNN with its embedding pre-trained by TransE~\citep{TransE} (Other knowledge embedding models can also provide pre-trained embeddings for S-GNN), and named the initialized features as $\bm{e}^0$.

In order to fully transfer the structure and knowledge information among entities in $\mathcal{G}_m$, S-GNN consists of several hidden layers to aggregate information following the structure of $\mathcal{G}_m$. At the $i$-th layer, given an entity $e \in \mathcal{E}_m$, S-GNN aggregates all information from its neighbors entity ${n}$ and ${r}$ in $\mathcal{G}_m$,
\begin{equation}
\bm{h}_{{n} \rightarrow e}^i = \left\{
        \begin{array}{c}
            \mathbf{W}^i[\bm{{n}} + \bm{r}; \bm{{n}}^{i-1}], ({n}, r, e) \in \mathcal{G}_m    \\
            \mathbf{W}^i[\bm{{n}} - \bm{r}; \bm{{n}}^{i-1}], (e, r, {n}) \in \mathcal{G}_m
        \end{array},
\right.
\end{equation}
where $\bm{{n}}^{i-1}$ is the embedding of ${n}$ at the $i-1$ layer, $\bm{{n}}$ and $\bm{r}$ are the entity and relation embeddings respectively pre-trained by TransE, $\bm{W}^i$ is a learnable linear matrix, and $[\cdot ; \cdot]$ denotes the horizontal concatenation of vectors. Then the embedding of $e$ at the $i$-th layer can be computed as
\begin{equation}
\label{eq:sgnn}
\bm{e}^i = f^i(\{ \bm{h}_{{n} \rightarrow e}^i \}_{{n} \in \mathcal{N}_{e}}),
\end{equation}
where $\mathcal{N}_{e}$ is the neighboring set of $e$, $f^i(\cdot)$ is the function to aggregate information at the $i$-th layer and will be introduced in detail next.

As not all information in the raw knowledge context $\mathcal{G}_m$ is useful for understanding the input text~{tokens} $S=\{w_j\}_{j=1}^N$, we design a special semantic attention mechanism as the function $f^i$ in Eq.~(\ref{eq:sgnn}) to filter out irrelevant information and aggregate essential information. The attention mechanism function $f$ can be formally denoted as follows,
\begin{equation}
\label{eq:attention}
\begin{aligned}
    f^i(\{ & \bm{h}_{\hat{e} \rightarrow e}^i \}_{\hat{e} \in \mathcal{N}_{e}}) = \\ \sum_{\hat{e}\in\mathcal{N}_{e}}
    &\frac{\exp(\bm{k}^{\top}_{\hat{e}}\bm{q})}{\sum_{\tilde{e}\in\mathcal{N}_{e}}\exp(\bm{k}^{\top}_{\tilde{e}}\bm{q})} \bm{h}_{\hat{e} \rightarrow e}^i,
\end{aligned}
\end{equation}
where $\bm{q}$, $\bm{k}_{{n}}$ are referred to as query and key vectors respectively. 

To dynamically select information according to textual context, the query vector $\bm{q}$ comes from the embedding of the input text~(tokens):
\begin{equation}
    \bm{q} = \sigma\left(\widehat{\bm{W}}^i \bm{s} + \widehat{\bm{b}}^i\right),
\end{equation}
where $\sigma=\tanh(\cdot)$, $\widehat{\mathbf{W}}^i$ and $\widehat{\mathbf{b}}^i$ are the learnable linear matrix and bias vector respectively for the query vector at the $i$-th layer, $\bm{s}$ is the whole semantic embedding of the input text~(tokens). Specially, following BERT~\cite{devlin2018bert}, we place a special token \texttt{[CLS]} at the beginning of the input sequence, and $\bm{s}$ is the output embedding of \texttt{[CLS]} computed by Eq.~(\ref{eqn:textual-encoder}).

The key vector $\bm{k}_{{n}}$ is based on the embedding of the relation between the entity $e$ and its neighboring entity ${n}$, and computed as
\begin{equation}
\begin{aligned}
\bm{k}_{{n}} =
\left\{\begin{matrix}
  & \widetilde{\bm{W}}^i \bm{(-r)} + \widetilde{\bm{b}}^i, &(e,r,{n}) \in \mathcal{G}_m \\
  & \widetilde{\bm{W}}^i \bm{r} + \widetilde{\bm{b}}^i, &({n},r,e) \in \mathcal{G}_m, 
\end{matrix}\right.
\end{aligned}
\end{equation}
where $\widetilde{\mathbf{W}}^i$ and $\widetilde{\mathbf{b}}^i$ are the learnable linear matrix and bias vector respectively for the key vector at the $i$-th layer. Two triples with head an tail entities switched will get the reverse key vectors.

In summary, S-GNN utilizes textual context to adjust the weight of feature aggregation, and finally selects and embeds knowledge related to the textual context into embbedings for PLMs. Hence, given the mentioned entity $m$, the output embedding of $m$ at the last layer of S-GNN is its final embedding computed by its dynamic knowledge context. For simplicity, given the input text~(tokens) $\{w_j\}_{j=1}^{N}$ and the mentioned entities $\{e_j\}_{j=1}^M$, the whole computation to achieve dynamic knowledge context embeddings is denoted as,
\begin{equation}
\{\bm{e}_{j}\}_{j=1}^{M} = \texttt{DK-Encoder}(\{e_{j}\}_{j=1}^{M}, \{w_j\}_{j=1}^{N}). 
\end{equation}

\subsection{Knowledge Fusion Encoder}

Knowledge fusion encoder aims to fuse the information of contextual entity embedding $\{\bm{e}_j\}_{j=1}^M$ and the text~(tokens) embedding $\{\bm{w}_j\}_{j=1}^N$. We leverage the encoder $\texttt{K-Encoder}(\cdot)$ similar to ~\citep{zhang2019ernie} to serve the purpose,
\begin{equation}
\begin{aligned}
    \{\bm{w}^o_j\}_{j=1}^N,\{\bm{e}^o_j\}_{j=1}^M& = \\
    \texttt{K-Encoder}( &\{\bm{w}_j\}_{j=1}^N,\{\bm{e}_j\}_{j=1}^M)
\end{aligned}
\end{equation}
We refer the readers to~\citep{zhang2019ernie} for more details. Roughly speaking, $\texttt{K-Encoder}(\cdot)$ consists of $P$ aggregators. As shown in Figure~\ref{fig:methodology}, in each aggregator, there are two multi-head self-attentions injecting text~(tokens) and contextual knowledge embeddings respectively, and a multi-layer perceptron~(MLP) fusing two heterogeneous features. 


\subsection{Training Details}

\paragraph{Pre-Training Strategies}
\label{Pre-training}

To incorporate knowledge embeddings into language understanding, we randomly mask token-entity alignments and let the model learn to predict all corresponding entities for these tokens by masking their alignments. We refer this to a denoising entity auto-encoder~(dEA), which is one of the pre-training tasks for existing knowledge-enhanced PLMs~\cite{zhang2019ernie}.

Besides, we choose BERT\BASESIZE~\cite{devlin2018bert}, RoBERTa\BASESIZE~\cite{liu2019RoBERTa}, and RoBERTa\LARGESIZE~\cite{liu2019RoBERTa} as our base models. Considering that our base models are originally pre-trained by different pre-training tasks, we have two different training objectives for them. 

For the \CokeBERTBASE, which is based on BERT\BASESIZE, the training objective can be described as:
\begin{equation}
\mathcal{L} = \mathcal{L}_{ \texttt{MLM}} + \mathcal{L}_{ \texttt{NSP}} + \mathcal{L}_{ \texttt{dEA}},
\end{equation}
where the $\mathcal{L}_{ \texttt{MLM}}$ and $\mathcal{L}_{ \texttt{NSP}}$ are loss functions for masked language model and next sentence prediction correspondingly. The denoising entity auto-encoder (dEA) loss is $\mathcal{L}_{\texttt{dEA}}$.


For \CokeRoBERTaBASE and \CokeRoBERTaLARGE, which are representatively based on RoBERTa\BASESIZE and RoBERTa\LARGESIZE, their training objective can be described as:
\begin{equation}
\mathcal{L} = \mathcal{L}_{\texttt{MLM}} + \mathcal{L}_{\texttt{dEA}},
\end{equation}
where the sentence prediction loss is removed.


\paragraph{Fine-Tuning for Downstream Tasks}
\label{Fine-tuning for Specific Tasks}
\Coke applies the fine-tuning procedure similar to BERT and take the final output embedding of the first token \texttt{[CLS]} for various common NLP tasks. Simliar to the previous knowledge-enhanced PLMs, for knowledge-driven tasks such as entity typing and relation classification, we apply specific fine-tuning procedures. As shown in Figure ~\ref{fig:methodology}, to help \Coke combine context information and entity mention attentively, we modify the input sequence with the mention markers. We attend the token which is in front of the entity mention as \texttt{[ENT]} and then use the final output embedding of \texttt{[ENT]} for the entity typing task. As for the relation classification task, we insert \texttt{[HD]} and \texttt{[TL]} tokens for head entities and tail entities respectively, and concatenate the \texttt{[HD]} representation and \texttt{[TL]} representation as final representation~\cite{baldini-soares-etal-2019-matching} for the task.

\section{Experiments} 
\label{Experiments}
In the experiments, we first introduce the training dataset and other training details of our model. After that, we give an empirical analysis to show the usefulness of the selected knowledge context. Then we compare Coke with several strong baselines in two typical knowledge-guided tasks including entity typing and relation classification. Finally, we perform an ablation study to show the effectiveness of our dynamic knowledge context encoder.

\subsection{Training Dataset}
\label{Training Dataset}
We use English Wikipedia\footnote{\url{https://en.wikipedia.org/}} as our pre-training corpus and align the entity mentions to Wikidata with widely-used entity linking tool TAGME~\cite{TAGME}. There are nearly $4,500$M subwords and $140$M entities in the pre-training corpus and we we sample $24,267,796$ fact triples, including $5,040,986$ entities in Wikidata. We conduct our experiments on the following datasets: FIGER, Open Entity, FewRel, and TACRED. The statistics of these datasets are shown in Table~\ref{statistic}. Besides, we use knowledge embeddings of WikiData released by~\cite{zhang2019ernie}.

\begin{table}[t]
\begin{center} 
\scalebox{0.8}{
{
\begin{tabular}{l|rrrrr}
\toprule
{\bf Dataset} & {\bf Train} & {\bf Dev} & {\bf Test} & {\bf Type} & {\bf Rel} \\ 
\midrule
FIGER & 2,000,000 & 10,000 & 563 & 113 & - \\
Open Entity & 2,000 & 2,000 & 2,000 & 6 & - \\
FewRel & 8,000 & 16,000 & 16,000 & - & 80 \\
TACRED & 68,124& 22,631 & 15,509 & - & 42 \\
\bottomrule
\end{tabular}}
}
\end{center} 
\caption{\label{statistic} The statistics of FIGER, Open Entity, FewRel, and TACRED datasets. }
\end{table}

\subsection{Experimental Settings}
\label{Pre-training Settings}
\paragraph{Training and Parameter Settings}
In experiments, we choose BERT\BASESIZE~\cite{devlin2018bert}, RoBERTa\BASESIZE and RoBERTa\LARGESIZE~\cite{liu2019RoBERTa} as our base models. To reduce the cost of training from scratch, we adopt these models' released parameters to initialize our text encoder and the rest of parameters of \Coke are all initialized randomly. 




For optimization, we set the learning rate as $5\times10^{-5}$, the max sequence length as $256$, the batch size as $32$, and the rest settings largely following the original PLMs. For fine-tuning, we use the same parameters as pre-training except the batch sizes and the learning rates. In all downstream tasks, we select the batch size from \{$16$,$32$,$64$\}, the learning rate is $2\times10^{-5}$, the number of epochs from \{$5$,$6$,$7$,$8$,$9$,$10$\}. The following ranges of value all perform well. Besides, to prevent \Coke from overfitting in FIGER, we use large batch size $1024$. We refer more details of training and hyper-parameter settings to our Appendix.

\paragraph{Baselines}
\label{Baselines}
We split baseline models into three groups: 

BERT\BASESIZE based models, RoBERTa\BASESIZE based models, and RoBERTa\LARGESIZE based models. For the sake of fairness, all models only incorporate factual knowledge from Wikidata. For knowledge-enhanced PLMs like ERNIE, KnowBert, and K-BERT, we re-implement them or use their released code for our experiments, and report the results which can match their results in the original papers. As \KADAPTER is similar to K-BERT and without any released code, we thus directly compare with K-BERT rather than \KADAPTER.

\begin{table*}[t]
\centering
\begin{minipage}[t]{\columnwidth}
\small
\centering
{ 
\setlength{\tabcolsep}{0.8mm}{ 
\scalebox{0.9}{
\begin{tabular}{l|ll}
\toprule 
   \multicolumn{3}{l}{\textbf{Text}: \textcolor{red}{\texttt{[CLS]}} Both Microsoft co-founder \textcolor{blue}{Bill Gates} and Facebook} \\
   \multicolumn{3}{l}{ co-founder \textcolor{blue}{Mark Zuckerberg} dropped out of Harvard and began} \\
   \multicolumn{3}{l}{ building their companies right around the same time.} \\
   \multicolumn{3}{l}{\textbf{Factual triple}: \textcolor{blue}{Mark Zuckerberg}, \textcolor{blue}{Bill Gates}, \textcolor{red}{alumnus}} \\
   \midrule\midrule
   \multicolumn{3}{c}{\textbf{Entity$_{h}$}: Mark Zuckerberg} \\
   \midrule
   \multicolumn{1}{c|}{\textbf{Importance}} & \multicolumn{1}{c}{\textbf{Entity$_{t}$}} &  \multicolumn{1}{c}{\textbf{Relation}}  \\
\rowcolor{GREEN_YL_1}
\multicolumn{1}{c|}{19\%} & \multicolumn{1}{c}{Harvard University} & \multicolumn{1}{c}{educated at} \\
\rowcolor{GREEN_YL_1}
\multicolumn{1}{c|}{19\%} & \multicolumn{1}{c}{Phillips Exeter Academy} & \multicolumn{1}{c}{educated at} \\
\rowcolor{GREEN_YL_1}
\multicolumn{1}{c|}{19\%} & \multicolumn{1}{c}{Ardsley High School} & \multicolumn{1}{c}{educated at} \\
\rowcolor{GREEN_YL_3}
\multicolumn{1}{c|}{10\%} & \multicolumn{1}{c}{Facebook} & \multicolumn{1}{c}{CEO of} \\
\rowcolor{GREEN_YL_3}
\multicolumn{1}{c|}{10\%} & \multicolumn{1}{c}{Chief executive officer} & \multicolumn{1}{c}{position held} \\
\rowcolor{GREEN_YL_5}
\multicolumn{1}{c|}{6\%} & \multicolumn{1}{c}{Businessperson} & \multicolumn{1}{c}{occupation} \\
\rowcolor{GREEN_YL_5}
\multicolumn{1}{c|}{6\%} & \multicolumn{1}{c}{Computer scientist} & \multicolumn{1}{c}{occupation} \\
\rowcolor{GREEN_YL_5}
\multicolumn{1}{c|}{6\%} & \multicolumn{1}{c}{Palo Alto, California} & \multicolumn{1}{c}{residence} \\
\rowcolor{GREEN_YL_6}
\multicolumn{1}{c|}{3\%} & \multicolumn{1}{c}{White Plains, New York } & \multicolumn{1}{c}{place of birth} \\
\rowcolor{GREEN_YL_7}
\multicolumn{1}{c|}{2\%} & \multicolumn{1}{c}{Mandarin Chinese} & \multicolumn{1}{c}{languages spoken} \\
\midrule
    \multicolumn{3}{c}{\textbf{Entity$_{h}$}: Bill Gates} \\
    \midrule
    \multicolumn{1}{c|}{\textbf{Importance}} & \multicolumn{1}{c}{\textbf{Entity$_{t}$}} &  \multicolumn{1}{c}{\textbf{Relation}}  \\
\rowcolor{GREEN_YL_1}
\multicolumn{1}{c|}{35\%} & \multicolumn{1}{c}{Harvard University} & \multicolumn{1}{c}{educated at} \\
\rowcolor{GREEN_YL_3}
\multicolumn{1}{c|}{11\%} & \multicolumn{1}{c}{Microsoft} & \multicolumn{1}{c}{CEO of} \\
\rowcolor{GREEN_YL_3}
\multicolumn{1}{c|}{11\%} & \multicolumn{1}{c}{Chief executive officer} & \multicolumn{1}{c}{position held} \\
\rowcolor{GREEN_YL_4}
\multicolumn{1}{c|}{9\%} & \multicolumn{1}{c}{American Academy of} & \multicolumn{1}{c}{member of} \\
\rowcolor{GREEN_YL_4}
\multicolumn{1}{c|}{} & \multicolumn{1}{c}{Arts and Sciences} & \multicolumn{1}{c}{} \\
\rowcolor{GREEN_YL_4}
\multicolumn{1}{c|}{9\%} & \multicolumn{1}{c}{National Academy} & \multicolumn{1}{c}{member of} \\
\rowcolor{GREEN_YL_4}
\multicolumn{1}{c|}{} & \multicolumn{1}{c}{of Engineering} & \multicolumn{1}{c}{} \\
\rowcolor{GREEN_YL_5}
\multicolumn{1}{c|}{6\%} & \multicolumn{1}{c}{Computer scientist} & \multicolumn{1}{c}{occupation} \\
\rowcolor{GREEN_YL_5}
\multicolumn{1}{c|}{6\%} & \multicolumn{1}{c}{Investor} & \multicolumn{1}{c}{occupation} \\
\rowcolor{GREEN_YL_5}
\multicolumn{1}{c|}{6\%} & \multicolumn{1}{c}{Businessperson} & \multicolumn{1}{c}{occupation} \\
\rowcolor{GREEN_YL_6}
\multicolumn{1}{c|}{4\%} & \multicolumn{1}{c}{Bill\&Melinda Gates Foundation} & \multicolumn{1}{c}{foundation of} \\
\rowcolor{GREEN_YL_7}
\multicolumn{1}{c|}{3\%} & \multicolumn{1}{c}{United States} & \multicolumn{1}{c}{citizenship} \\
\bottomrule 
\multicolumn{3}{c}{} \\ 
\end{tabular}}}}
\end{minipage}
\quad\ 
\begin{minipage}[t]{\columnwidth} 
\small
\centering
{ 
\setlength{\tabcolsep}{0.8mm}{
\scalebox{0.9}{
\begin{tabular}{l|ll}
\toprule 
   \multicolumn{3}{l}{\textbf{Text}: \textcolor{red}{\texttt{[CLS]}} \textcolor{blue}{Bill Gates} and \textcolor{blue}{Mark Zuckerberg} are working together} \\
   \multicolumn{3}{l}{to fund research for COVID-19 treatments.} \\
   \multicolumn{3}{l}{\textbf{Factual triple}: \textcolor{blue}{Mark Zuckerberg}, \textcolor{blue}{Bill Gates}, \textcolor{red}{cooperate}} \\
   \midrule\midrule
   \multicolumn{3}{c}{\textbf{Entity$_{h}$}: Mark Zuckerberg} \\
   \midrule
   \multicolumn{1}{c|}{\textbf{Importance}} & \multicolumn{1}{c}{\textbf{Entity$_{t}$}} &  \multicolumn{1}{c}{\textbf{Relation}}  \\
\rowcolor{GREEN_YL_1}
\multicolumn{1}{c|}{15\%} & \multicolumn{1}{c}{Facebook} & \multicolumn{1}{c}{CEO of} \\
\rowcolor{GREEN_YL_2}
\multicolumn{1}{c|}{14\%} & \multicolumn{1}{c}{Chief executive officer} & \multicolumn{1}{c}{position held} \\
\rowcolor{GREEN_YL_3}
\multicolumn{1}{c|}{11\%} & \multicolumn{1}{c}{Businessperson} & \multicolumn{1}{c}{occupation} \\
\rowcolor{GREEN_YL_3}
\multicolumn{1}{c|}{11\%} & \multicolumn{1}{c}{Computer scientist} & \multicolumn{1}{c}{occupation} \\
\rowcolor{GREEN_YL_5}
\multicolumn{1}{c|}{9\%} & \multicolumn{1}{c}{Harvard University} & \multicolumn{1}{c}{educated at} \\
\rowcolor{GREEN_YL_5}
\multicolumn{1}{c|}{9\%} & \multicolumn{1}{c}{Phillips Exeter Academy} & \multicolumn{1}{c}{educated at} \\
\rowcolor{GREEN_YL_5}
\multicolumn{1}{c|}{9\%} & \multicolumn{1}{c}{Ardsley High School} & \multicolumn{1}{c}{educated at} \\
\rowcolor{GREEN_YL_6}
\multicolumn{1}{c|}{8\%} & \multicolumn{1}{c}{Palo Alto, California} & \multicolumn{1}{c}{residence} \\
\rowcolor{GREEN_YL_7}
\multicolumn{1}{c|}{7\%} & \multicolumn{1}{c}{White Plains, New York} & \multicolumn{1}{c}{place of birth} \\
\rowcolor{GREEN_YL_7}
\multicolumn{1}{c|}{7\%} & \multicolumn{1}{c}{Mandarin Chinese} & \multicolumn{1}{c}{languages spoken} \\
\midrule
    \multicolumn{3}{c}{\textbf{Entity$_{h}$}: Bill Gates} \\
    \midrule
    \multicolumn{1}{c|}{\textbf{Importance}} & \multicolumn{1}{c}{\textbf{Entity$_{t}$}} &  \multicolumn{1}{c}{\textbf{Relation}}  \\
\rowcolor{GREEN_YL_1}
\multicolumn{1}{c|}{33\%} & \multicolumn{1}{c}{Bill\&Melinda Gates Foundation} & \multicolumn{1}{c}{foundation of} \\
\rowcolor{GREEN_YL_3}
\multicolumn{1}{c|}{10\%} & \multicolumn{1}{c}{Microsoft} & \multicolumn{1}{c}{CEO of} \\
\rowcolor{GREEN_YL_4}
\multicolumn{1}{c|}{9\%} & \multicolumn{1}{c}{Chief executive officer} & \multicolumn{1}{c}{position held} \\
\rowcolor{GREEN_YL_5}
\multicolumn{1}{c|}{8\%} & \multicolumn{1}{c}{American Academy of} & \multicolumn{1}{c}{member of} \\
\rowcolor{GREEN_YL_5}
\multicolumn{1}{c|}{} & \multicolumn{1}{c}{Arts and Sciences} & \multicolumn{1}{c}{} \\
\rowcolor{GREEN_YL_5}
\multicolumn{1}{c|}{8\%} & \multicolumn{1}{c}{National Academy} & \multicolumn{1}{c}{member of} \\
\rowcolor{GREEN_YL_5}
\multicolumn{1}{c|}{} & \multicolumn{1}{c}{of Engineering} & \multicolumn{1}{c}{} \\
\rowcolor{GREEN_YL_6}
\multicolumn{1}{c|}{7\%} & \multicolumn{1}{c}{Computer scientist} & \multicolumn{1}{c}{occupation} \\
\rowcolor{GREEN_YL_6}
\multicolumn{1}{c|}{7\%} & \multicolumn{1}{c}{Investor} & \multicolumn{1}{c}{occupation} \\
\rowcolor{GREEN_YL_6}
\multicolumn{1}{c|}{7\%} & \multicolumn{1}{c}{Businessperson} & \multicolumn{1}{c}{occupation} \\
\rowcolor{GREEN_YL_7}
\multicolumn{1}{c|}{6\%} & \multicolumn{1}{c}{Harvard University } & \multicolumn{1}{c}{educated at} \\
\rowcolor{GREEN_YL_7}
\multicolumn{1}{c|}{5\%} & \multicolumn{1}{c}{United States} & \multicolumn{1}{c}{citizenship} \\
\bottomrule 
\end{tabular}}}}
\end{minipage}
\caption{The shade of color expresses the importance of triples for a given sentence.}
\label{case_study}
\end{table*}

\subsection{Empirical Analysis for Dynamically Selecting Knowledge Context}
\label{Empirical Analysis}
To demonstrate \Coke is able to capture useful information from KGs, we design a qualitative and quantitative experiments to evaluate \Coke.

In the qualitative experiment, given the same entity mentions in different context, we adopt PLMs for selecting text-related $1$-hop triples~(``$1$-hop knowledge context'') from Wikidata, which is similar to Eq.~(\ref{eq:attention}) without summation. More specifically, we apply the \texttt{[CLS]} of the input text~(tokens) computed by these PLMs to attend each neighbouring triple of entity mentions.

As shown in Table~\ref{case_study}, when given the sentence ``\textsl{$\ldots$ Bill Gates and Mark Zuckerberg dropped out of Harvard $\ldots$}'' indicating the relation alumni between Mark Zuckerberg and Bill Gates, our model pays more attention to the factual knowledge of their education. Yet when given the sentence ``\textsl{Bill Gates and Mark Zuckerberg are working together $\ldots$}'' indicating the cooperation between Mark Zuckerberg and Bill Gates, the factual knowledge of their enterprises is considered by our model. Apparently, we can find the importance scores of attended triples is interpretable and can help us understand the semantics more clearly.


In the quantitative experiment, we annotate the test sets of FewRel and TACRED. Given a sample, including context and the corresponding entity mentions, we manually annotate its $1$-hop triples by judging the relevance between context and triples. Finally, we extract $15981$ instances from FewRel and $5684$ instances from TACRED. By ranking importance scores of all triples for an entity mention and setting a threshold, we can obtain positive triples and negative triples to calculate F1 scores for evaluation. 

To fairly demonstrate effectiveness of extracting triples via \Coke, we choose ERNIE as our baseline model, which inherently aligns the language embedding space and KG embedding space using the same training data as \Coke. As shown in Table~\ref{tab:caseres}, the F1 scores of \Coke are better than the baseline model by $14.8$\%-$17.8\%$ on FewRel and $14.5$\%-$18.3$\% on TACRED. 

\begin{table}[t]
\begin{center} 
\scalebox{0.95}{
\scalebox{1.0}{ 
\setlength{\tabcolsep}{3pt}{ 
\begin{tabular}{l|lll|lll}
\toprule
  & \multicolumn{3}{c|}{\textbf{FewRel}} & \multicolumn{3}{c}{\textbf{TACRED}}  \\ 
  & \multicolumn{1}{c}{\textbf{P}} & \multicolumn{1}{c}{\textbf{R}} & \multicolumn{1}{c|}{\textbf{F1}} & \multicolumn{1}{c}{\textbf{P}} & \multicolumn{1}{c}{\textbf{R}} & \multicolumn{1}{c}{\textbf{F1}}  \\
\midrule\midrule
ERNIE & \multicolumn{1}{c}{87.6} & \multicolumn{1}{c}{50.6} & \multicolumn{1}{c|}{64.1} & \multicolumn{1}{c}{81.1} & \multicolumn{1}{c}{41.8} & \multicolumn{1}{c}{55.1}   \\
\midrule
\CokeBERTBASE & \multicolumn{1}{c}{\textbf{87.9}} & \multicolumn{1}{c}{71.5} & \multicolumn{1}{c|}{78.9} & \multicolumn{1}{c}{\textbf{86.1}} & \multicolumn{1}{c}{58.4} & \multicolumn{1}{c}{69.6}   \\
\CokeRoBERTaBASE & \multicolumn{1}{c}{79.8} & \multicolumn{1}{c}{\textbf{84.0}} & \multicolumn{1}{c|}{\textbf{81.9}} & \multicolumn{1}{c}{74.9} & \multicolumn{1}{c}{\textbf{72.0}} & \multicolumn{1}{c}{\textbf{73.4}} \\
\bottomrule
\end{tabular}}}}
\end{center}  
\caption{\label{tab:caseres} The results of capturing positive triples from the labeled triples on FewRel and TACRED (\%). }
\end{table}

\begin{table*}[t]
\begin{center} 
\small
{
\setlength{\tabcolsep}{7pt}{
\begin{tabular}{l|llllll|llllll}
\toprule
  \textbf{Task} & \multicolumn{6}{c|}{\textbf{Entity Typing }} & \multicolumn{6}{c}{\textbf{Relation Classification}} \\
  \midrule
  \textbf{Dataset} & \multicolumn{3}{c|}{\textbf{Open Entity}} & \multicolumn{3}{c|}{\textbf{FIGER}}  & \multicolumn{3}{c|}{\textbf{FewRel}} & \multicolumn{3}{c}{\textbf{TACRED}}
  \\ 
  \textbf{Metric} & \multicolumn{1}{c}{\textbf{P}} & \multicolumn{1}{c}{\textbf{R}} & \multicolumn{1}{c|}{\textbf{F1}} & \multicolumn{1}{c}{\textbf{Acc.}} & \multicolumn{1}{c}{\textbf{Macro}} & \multicolumn{1}{c|}{\textbf{Micro}}  & \multicolumn{1}{c}{\textbf{P}} & \multicolumn{1}{c}{\textbf{R}} & \multicolumn{1}{c|}{\textbf{F1}} & \multicolumn{1}{c}{\textbf{P}} & \multicolumn{1}{c}{\textbf{R}} & \multicolumn{1}{c}{\textbf{F1}} \\
\midrule\midrule
\multicolumn{13}{c}{Pre-Trained Language Models}\\
\midrule
BERT\BASESIZE & \multicolumn{1}{c}{76.2} & \multicolumn{1}{c}{71.0} & \multicolumn{1}{c|}{73.6} & \multicolumn{1}{c}{52.0} & \multicolumn{1}{c}{75.2} & \multicolumn{1}{c|}{71.6} & \multicolumn{1}{c}{85.0} & \multicolumn{1}{c}{85.1} & \multicolumn{1}{c|}{84.9} & \multicolumn{1}{c}{67.2} & \multicolumn{1}{c}{64.8} & \multicolumn{1}{c}{66.0}  \\
RoBERTa\BASESIZE & \multicolumn{1}{c}{75.3} & \multicolumn{1}{c}{73.2} & \multicolumn{1}{c|}{74.2} & \multicolumn{1}{c}{56.3} & \multicolumn{1}{c}{76.9} & \multicolumn{1}{c|}{74.2} & \multicolumn{1}{c}{86.3} & \multicolumn{1}{c}{86.3} & \multicolumn{1}{c|}{86.3} & \multicolumn{1}{c}{73.0} & \multicolumn{1}{c}{68.7} & \multicolumn{1}{c}{70.8}  \\
RoBERTa\LARGESIZE & \multicolumn{1}{c}{78.5} & \multicolumn{1}{c}{72.7} & \multicolumn{1}{c|}{75.5} & \multicolumn{1}{c}{57.1} & \multicolumn{1}{c}{\textbf{82.4}} & \multicolumn{1}{c|}{76.5} & \multicolumn{1}{c}{88.4} & \multicolumn{1}{c}{88.4} & \multicolumn{1}{c|}{88.4} & \multicolumn{1}{c}{\textbf{74.3}} & \multicolumn{1}{c}{66.8} & \multicolumn{1}{c}{70.4}   \\
\midrule\midrule
\multicolumn{13}{c}{Knowledge Enhance Pre-Trained Language Models}\\
\midrule
ERNIE & \multicolumn{1}{c}{78.4} & \multicolumn{1}{c}{72.9} & \multicolumn{1}{c|}{75.6} & \multicolumn{1}{c}{57.2} & \multicolumn{1}{c}{76.5} & \multicolumn{1}{c|}{73.4} & \multicolumn{1}{c}{88.5} & \multicolumn{1}{c}{88.4} & \multicolumn{1}{c|}{88.3} & \multicolumn{1}{c}{69.9} & \multicolumn{1}{c}{66.0} & \multicolumn{1}{c}{67.9} \\
K-BERT & \multicolumn{1}{c}{76.7} & \multicolumn{1}{c}{71.5} & \multicolumn{1}{c|}{74.0} & \multicolumn{1}{c}{56.5} & \multicolumn{1}{c}{77.1} & \multicolumn{1}{c|}{73.8} & \multicolumn{1}{c}{83.1} & \multicolumn{1}{c}{85.9} & \multicolumn{1}{c|}{84.3} & \multicolumn{1}{c}{68.1} & \multicolumn{1}{c}{66.1} & \multicolumn{1}{c}{67.1} \\
KnowBert-Wiki & \multicolumn{1}{c}{\textbf{78.6}} & \multicolumn{1}{c}{71.6} & \multicolumn{1}{c|}{75.0} & \multicolumn{1}{c}{57.0} & \multicolumn{1}{c}{79.8} & \multicolumn{1}{c|}{75.0} & \multicolumn{1}{c}{89.2} & \multicolumn{1}{c}{89.2} & \multicolumn{1}{c|}{89.2} & \multicolumn{1}{c}{71.1} & \multicolumn{1}{c}{66.8} & \multicolumn{1}{c}{68.9} \\
\midrule\midrule
\multicolumn{13}{c}{Contextual Knowledge Enhanced Pre-Trained Language Models}\\
\midrule
\CokeBERTBASE & \multicolumn{1}{c}{78.0} & \multicolumn{1}{c}{73.3} & \multicolumn{1}{c|}{75.6} & \multicolumn{1}{c}{57.9} & \multicolumn{1}{c}{79.7} & \multicolumn{1}{c|}{75.3} & \multicolumn{1}{c}{89.4} & \multicolumn{1}{c}{89.4} & \multicolumn{1}{c|}{89.4} & \multicolumn{1}{c}{71.0} & \multicolumn{1}{c}{66.9} & \multicolumn{1}{c}{68.9}  \\
\CokeRoBERTaBASE & \multicolumn{1}{c}{76.8} & \multicolumn{1}{c}{74.2} & \multicolumn{1}{c|}{75.6} & \multicolumn{1}{c}{\textbf{62.2}} & \multicolumn{1}{c}{82.3} & \multicolumn{1}{c|}{77.7} & \multicolumn{1}{c}{90.1} & \multicolumn{1}{c}{90.1} & \multicolumn{1}{c|}{90.1} & \multicolumn{1}{c}{71.3} & \multicolumn{1}{c}{71.0} & \multicolumn{1}{c}{\textbf{71.1}} \\
\CokeRoBERTaLARGE & \multicolumn{1}{c}{75.3} & \multicolumn{1}{c}{\textbf{76.2}} & \multicolumn{1}{c|}{\textbf{75.7}} & \multicolumn{1}{c}{58.3} & \multicolumn{1}{c}{82.3} & \multicolumn{1}{c|}{\textbf{77.8}} & \multicolumn{1}{c}{\textbf{91.1}} & \multicolumn{1}{c}{\textbf{91.1}} & \multicolumn{1}{c|}{\textbf{91.1}} & \multicolumn{1}{c}{69.9} & \multicolumn{1}{c}{\textbf{71.8}} & \multicolumn{1}{c}{70.8} \\
\bottomrule
\end{tabular}}}
\end{center}  
\caption{\label{Relation Classification and Entity Typing} The results of various models for Relation Classification and Entity Typing~(\%). }
\end{table*}

\subsection{Overall Evaluation Results}
\label{Overall Evaluations Section}
In this section, we compare our models with various effective PLMs on entity typing and relation classification, including both vanilla PLMs and knowledge-enhanced PLMs.

\paragraph{Entity Typing}
\label{Entity Typing Section}
Given an entity mention and its corresponding sentence, entity typing requires to classify the entity mention into its types. For this task, we fine-tune \Coke on FIGER~\cite{ling-etal-2015-design} and Open Entity~\cite{choi-etal-2018-ultra}. The training set of FIGER is labeled with distant supervision, and its test set is annotated by human. Open Entity is a completely manually-annotated dataset. We compare our model with baseline models we mentioned in Baselines~\ref{Baselines}. 

As shown in Table~\ref{Relation Classification and Entity Typing}, \Coke can achieve comparable F1 scores on Open Entity. On FIGER, \Coke significantly outperform the BERT\BASESIZE and RoBERTa\BASESIZE by $3.7$\% and $3.5$\% Micro scores respectively. Besides, the performance of \Coke  is better than other baseline models as well.
It directly demonstrates that \Coke has better ability to reduce the noisy label challenge in FIGER than the baseline models that we mentioned above. 

Moreover, we found the domain of FIGER is similar to Wikidata, this is consistent with the observation in the empirical analysis section, which further highlights the importance of selecting knowledge context cross domains.


\paragraph{Relation Classification}
\label{Relation Extraction Section}
Relation classification aims to determine the correct relation between two entities in a given sentence. We fine-tune \Coke on two widely-used benchmark dataset FewRel~\cite{han-etal-2018-fewrel} and TACRED~\cite{zhang2017tacred}. We also compare our model with baseline models we mentioned in Baselines~\ref{Baselines}. 

On FewRel, \Coke significantly outperforms the BERT\BASESIZE and RoBERTa\BASESIZE by $4.5$\% and $3.8$\% F1 scores respectively as shown in Table~\ref{Relation Classification and Entity Typing}. It directly demonstrates that \Coke can capture the relation between two entities better than ERNIE by considering the information of higher-order neighbours, especially in small dataset FewRel. 

Besides, \Coke models have comparable results with other baseline models on TACRED but achieve substantially improvements on FewRel. As we mentioned before, the domain of FewRel data is more similar to Wikidata and therefore it gains more benefit from pre-training.

\subsection{Ablation Study}
\label{Ablation Study Section}

In order to indicate the effect of S-GNN on the process of dynamically selecting knowledge context, we conduct essential ablation studies for different modules in S-GNN.

\paragraph{K-Hop Sub-Graphs}
\label{K-hop sub-graphs Section}
In this section, we explore the effects of  dynamic knowledge context encoder. There are two main components in the dynamic knowledge context encoder: raw knowledge context construction and S-GNN. \Coke applies raw knowledge context construction to sample $K$-hop sub-graphs, and then incorporates S-GNN to embed informative knowledge in the raw context. 

\begin{figure}[t]
\small
\subfigure[FewRel]{
\includegraphics[width=0.47\linewidth]{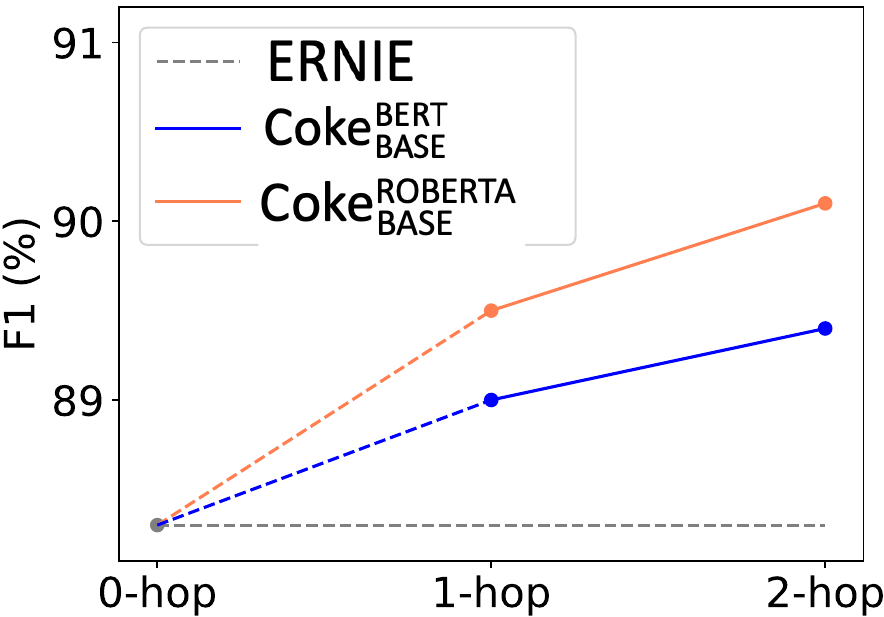}}
\subfigure[FIGER]{
\includegraphics[width=0.47\linewidth]{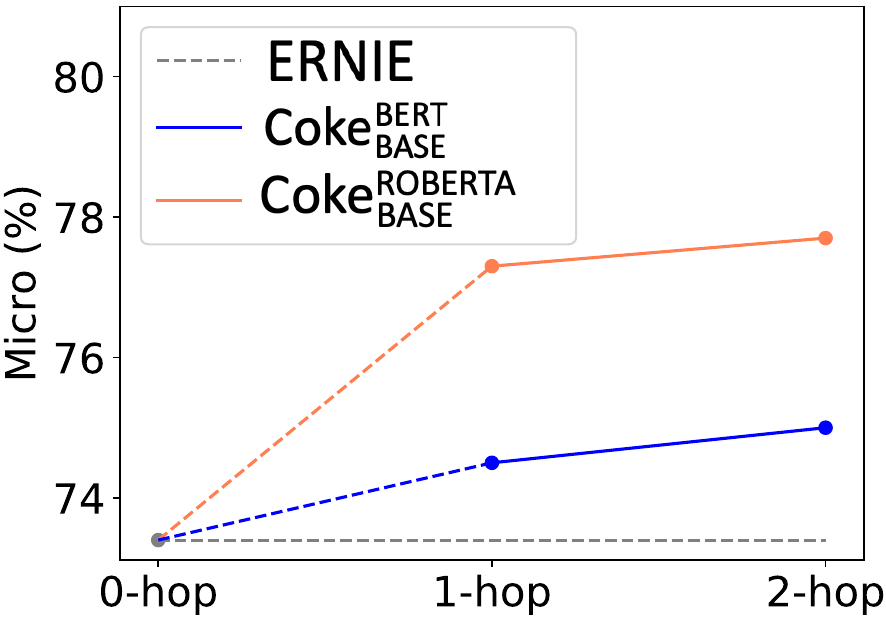}}
\caption{The results of \Coke incorporating $K$-hop sub-graphs (\%).}
\label{K-hop}
\vspace{-0.5em}
\end{figure}

From Figure~\ref{K-hop}, we find that \Coke incorporating the $2$-hop sub-graph outperforms by $0.4$\% to $0.6$\% than incorporating the $1$-hop sub-graph. It proves that considering a wider range of knowledge can lead to better entity embeddings.

\begin{figure}[t]
\small
\subfigure[FewRel]{
\includegraphics[width=0.47\linewidth]{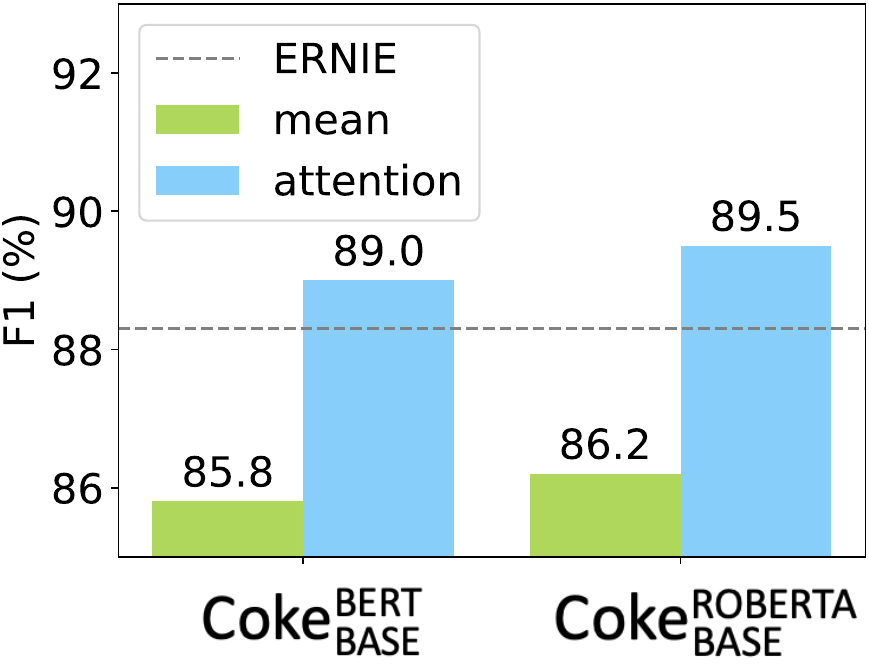}}
\subfigure[FIGER]{
\includegraphics[width=0.47\linewidth]{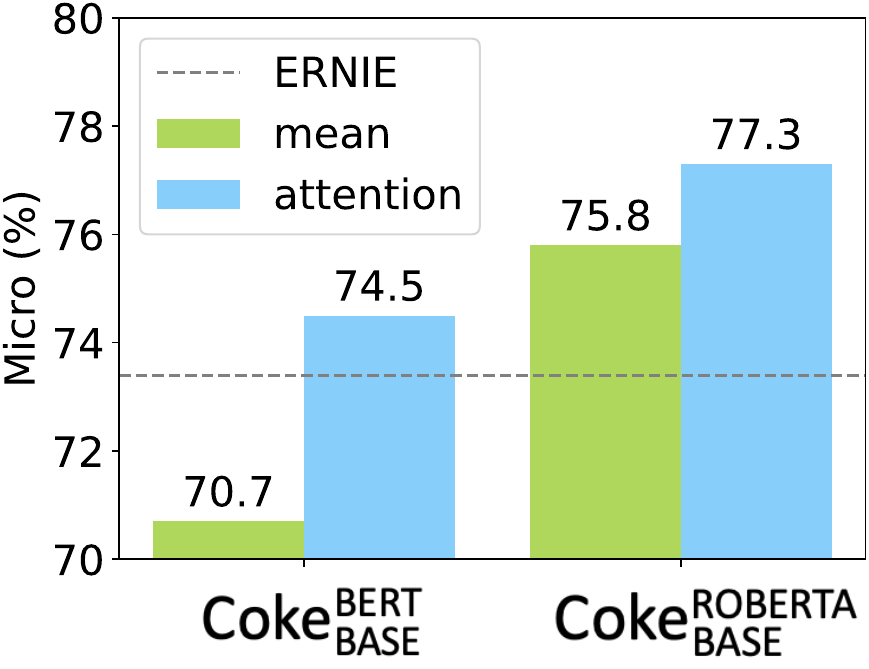}}
\caption{The effect of the attention mechanism and its simplified versions (\%). }
\label{attention-mean}
\vspace{-0.5em}
\end{figure}

\paragraph{Attention Mechanism}
\label{Attention Mechanism Section}
In S-GNN, there is an essential mechanism: attention. It takes responsibility for weighing how much knowledge matches the text and help compute final dynamic contextual embeddings. To further demonstrate the effect of the attention mechanism, we simplify it with a mean-pooling operation to aggregate features. From Figure~\ref{attention-mean}, we can find that the attention mechanism outperforms than the mean-pooling mechanism and fixed embeddings (ERNIE), indicating the effectiveness of our attention mechanism.

\section{Conclusion and Future Work}
We have proposed an effective and general framework to enable PLMs to dynamically select appropriate knowledge context with textual context, and then insert the embedded knowledge into PLMs.
The experiments demonstrate that \Coke can achieve comparable results with the state-of-the-art knowledge-enhanced PLMs in the entity typing and relation classification. \Coke dynamically selects knowledge context with textual context is more interpretable than injecting all knowledge context from KGs. In the empirical analysis, \Coke demonstrates the effective selection of knowledge context as well. This direction may lead to more general and effective language understanding. In the future, we will continue to explore how to inject other type of knowledge (e.g. linguistic knowledge) in conjunction with factual knowledge to further enhance PLMs. And it is also an interesting direction to explore how to continually inject emerging factual knowledge into PLMs without re-training the whole model.

\bibliography{emnlp2020}
\bibliographystyle{emnlp2020}

\end{document}